\documentclass[%
 preprint,
 amsmath,amssymb,
]{revtex4-1}

\usepackage{graphicx}
\usepackage{dcolumn}
\usepackage{bm}
\usepackage{hyperref}
\usepackage{amsmath}
\usepackage{multirow}
\usepackage{url}
\usepackage{subcaption}

\usepackage[utf8]{inputenc}
\usepackage{booktabs}
\usepackage{setspace}
\usepackage[table,xcdraw]{xcolor}

\usepackage{color}
\definecolor{redcolor}{rgb}{1.0,0.,0.}

\begin{document}

\preprint{}


\title{Using virtual edges to extract keywords from texts modeled as complex networks}
\author{Jorge A. V. Tohalino}%
\affiliation{%
Institute of Mathematics and Computer Science,  University of S\~{a}o Paulo,
S\~{a}o Carlos, SP,  Brazil
}%
\author{Thiago C. Silva}%
\affiliation{Universidade Católica de Brasília, Brasília, DF, Brazil}
\author{Diego R. Amancio}%
\email{diego@icmc.usp.br}
\affiliation{%
Institute of Mathematics and Computer Science, Department of Computer Science, University of S\~{a}o Paulo,
S\~{a}o Carlos, SP,  Brazil
}%


\date{\today}

\begin{abstract}
The keyword extraction task is an important NLP task in many text mining applications. Graph-based methods have been commonly used to automatically find the key concepts in texts, however, relevant information provided by embeddings has not been widely used to enrich the graph structure. Here we modeled texts co-occurrence networks, where nodes are words and edges are established either by contextual or semantical similarity. We compared two embedding approaches (Word2vec and BERT) to check whether edges created via word embeddings can improve the quality of the keyword extraction method. We found that, in fact, the use of virtual edges can improve the discriminability of co-occurrence networks. The best performance was obtained when we considered low percentages of addition of virtual (embedding) edges. A comparative analysis of structural and dynamical network metrics revealed the degree, PageRank, and accessibility are the metrics displaying the best performance in the model enriched with virtual edges.
\end{abstract}

\maketitle


\section{Introduction}\label{sec:intro}

In recent years, there has been a large increase in textual information available on the Internet. Examples include newspapers, social network comments, books, encyclopedias and scientific articles. In order to make sense and summarize such a large volume of data, several NLP applications have been proposed. One particular task is the keyword extraction task, which consists of selecting a set of words (or topics) that best represent the content of a document~\cite{vijaya2022graph}. Finding keywords in  multiple documents is important because manually finding the most central words can be an expensive and time-consuming task for human annotators. Since keyword extraction provides a compact representation of the document, many applications can benefit from this task: automatic indexing, automatic document summarization, automatic document classification, document clustering, automatic filtering, among other applications~\cite{bharti2017automatic, an2005keyword, hammouda2005corephrase}. 

Different approaches have been considered for the keyword extraction task~\cite{bharti2017automatic}. The simplest models are the statistical models that study the statistical information regarding the spatial use of words in each text as well as their frequency of use~\cite{herrera2008statistical}. These methods include for example the well-known Term Frequency (TF) or Term Frequency-Inverse Document Frequency (TF-IDF). Approaches based on linguistic and syntactic analysis have also been used to address this task~\cite{vega2019multi}. Additionally, several features extracted from the previous approaches can be  used in machine learning algorithms. The main goal of these methods is to detect keywords via binary classification~\cite{jiang2009ranking}.

Graph-based approaches have also been used to detect keywords~\cite{vega2019multi, Tohalino2018}. The objective of these methods is to represent each document as a network of words and then apply a set of centrality measurements to assign a relevance value for each network node. In this way, the most central nodes represent the automatic keywords found for each document. Most of these approaches have used word co-occurrence networks, where an edge exists between two words if they are adjacent. However, different strategies to connect words have not been extensively studied, with most of the works considering larger window contexts in the co-occurrence model~\cite{vega2019multi, mihalcea-tarau-2004-textrank}.

In this paper, we propose a graph-based method for keyword extraction, where texts are represented as co-occurrence networks and edges are established in a twofold manner. In addition to word adjacency models, we consider further contexts to connect words. In order to better represent the relationship between words, we also link words that do not necessarily co-occur in the text, but are semantically similar. 
Our motivation is to enrich the representation by including the so-called virtual edges. Consequently, hidden similarities are explicitly represented in the model. 
Our hypothesis is that the included virtual edges can be used to improve the traditional co-occurrence network representation based on word adjacency relationships alone. 
In the proposed model, the virtual edges were constructed from the word vectors generated by the Word2Vec and BERT embedding models~\cite{mikolov2013efficient,devlin2018bert}. After the networks are constructed, we computed the centrality values for each node (word) of the network. We used several structural and dynamical network measurements to identify the key concepts in texts. We also probed the effect of using the weighted versions of these measurements. 
The efficiency of our methods was evaluated in different datasets comprising documents of various sizes.

We have found several interesting results from this analysis. First, we observed that including virtual edges can improve the performance in retrieving keywords. The 
fraction of included virtual edges required to yield optimized results turned out to be relatively low. A negative performance effect was observed, however, when too many virtual edges were included.  
Concerning the embedding method, both considered strategies -- Word2vec and BERT -- yielded similar performance. The network metrics with the best performance were the degree, PageRank, and Accessibility. Surprisingly, when the weighted versions of the traditional metrics had a poor performance.
Our results reinforce the potential of enriching networks in multiple text network applications~\cite{stella2020cognitive,stella2017multiplex,castro2019multiplex}.

This manuscript is organized as follows: Section~\ref{sec:related} presents a summary of the related works for keyword extraction. The description of the datasets, as well as the proposed methodology, are described in Section~\ref{sec:methodology}. In this section, we describe the network creation stage and the process of extracting keywords using network centrality metrics. The results are discussed in Section~\ref{sec:results}. describes the obtained results and the analysis of each network measurement. Finally, in Section~\ref{sec:conclusions} we present the conclusions and perspectives for future works.

\section{Related works}\label{sec:related}

Studies addressing the keyword extraction problem can be grouped into three main approaches: statistical and network-based methods~\cite{merrouni2020automatic, hasan2014automatic}. The objective of statistical methods is to rank words using their statistical distribution along the text~\cite{herrera2008statistical,carretero2013improving}.  A very simple approach is described is the one relying on word frequency~\cite{luhn1958automatic}. According to this approach, words are sorted according to their frequency, and the most and less frequent words are disregarded. Such words are disregarded because they are common words (such as prepositions) or rare (not relevant) words.
Note that frequency-based approaches do not consider the structure of the text, since a shuffled, meaningless version of the same text would provide the same set of keywords. To overcome the weaknesses of  frequency-based methods, word clustering and word entropy methods were then proposed~\cite{ortuno2002keyword, herrera2008statistical}. Some modifications of these techniques include term-frequency inverse-document frequency approaches~\cite{qaiser2018text}.

In \cite{ortuno2002keyword}, the authors found that relevant words, which are more closely related to the main topics of the text, are generally concentrated in certain regions of the text. Keywords are usually unevenly distributed along the text and tend to form clusters. Conversely, common words are more regularly distributed along texts. 
A combination of both spatial clustering and frequency was proposed in~\cite{carpena2009level}. The authors  used the Shannon's entropy metric to define a method based on the information content of the sequence of occurrences of each word in the text. They used text partitions to calculate the entropy of all words. Because relevant words are unevenly distributed, the heterogeneity of word distribution in different partitions can be captured via entropy. In comparison to clustering-based methods,  an improved  performance was reported with entropy-based techniques. The main advantage of the statistical techniques is that they do not require any knowledge of the language and thus can be used any analyze even unknown documents~\cite{de2019paragraph}. 


Graph-based approaches include the representation of the relationship of words as networks~\cite{zhan2012keyword,grineva2009extracting,lahiri2014keyword}. In~\cite{grineva2009extracting}, the authors modeled documents as graphs of semantic relationships between the words. The weight linking two words modeled the semantic relatedness computed as measured via Wikipedia. The strategy considered that the words related to  central topics  tend to be grouped into densely connected  network communities, while common words are organized in weakly connected communities.  This method was found to be particularly  effective in removing noisy information.  
A similar study represented texts as word co-occurrence networks considering weighted and directed networks~\cite{lahiri2014keyword}. They used several centrality network measurements to find the relevant words. The authors concluded that network measurements can be successfully used for keyword extraction without the need for large external corpora. They also found that simpler centrality metrics like node degree or strength outperformed more complex and computationally expensive metrics in the proposed methodology. Related strategies have been used to find key concepts for the purpose of text summarization~\cite{Tohalino2018}.

Finally, word co-occurrence network considering larger co-occurrence contexts was proposed in~\cite{vega2019multi}. Different from other works, an edge is created if words co-occur within  a window comprising three words. Using a combination of feature selection and clustering methods to find the best set of keywords, the authors obtained optimized results in comparison to other works that only employed traditional co-occurrence networks. The authors also found that several network metrics are strongly correlated, yielding thus equivalent performance.

Differently from the previous works, here we propose a graph-based method for keyword extraction using \emph{enriched} word co-occurrence networks. In addition to the edges established via word adjacency, we considered edges created via embedding similarity. Several centrality measurements were then used to find the most relevant words. As we shall show, our approach outperforms both the traditional word adjacency model and its modified version considering larger contexts. 

\section{Material and methods}\label{sec:methodology}


The framework proposed to detect keywords via word embeddings and graph modeling comprises the following four main steps:  i) text pre-processing; ii) word vectorization; iii) network creation; and iv) word ranking and keyword extraction. 

\begin{enumerate}

    \item \emph{Pre-processing}: This phase comprises the required steps to conveniently pre-process the datasets. This step encompasses sentence segmentation, stopword removal and text stemming. In Section~\ref{sec:preprop}, we provide a brief description of the pre-processing steps we applied.    
    
    \item \emph{Word vectorization}: we considered the embeddings models to represent the words. The embeddings are important   for identifying similar words that do not co-occur in the text. Section~\ref{sec:methodology:embeddings} provides a detailed explanation of the word embedding methods used in this work.

    \item \emph{Network creation}: We modeled the documents as word co-occurrence networks, where nodes represent words and edges are established based on the neighbors of each word. We also considered ``virtual'' edges, which were generated based on the similarity between two words. The similarity is computed based on the word vectorization. This is an essential step for capturing long-range, semantical relationships. In Section~\ref{sec:methodology:network}, we explain the adopted methodology for network creation. 
    
    \item \emph{Keyword extraction}: We used several network centrality measurements to rank the words for each document. Such measurements are used to give an importance value or relevance weight to each node from the network. The top $N$ ranked words were considered keywords. Section~\ref{sec:methodology:ranking} describes the keyword extraction step.         
    
\end{enumerate}

\begin{figure}[h]
    \centering
    \includegraphics[width=1.0\textwidth]{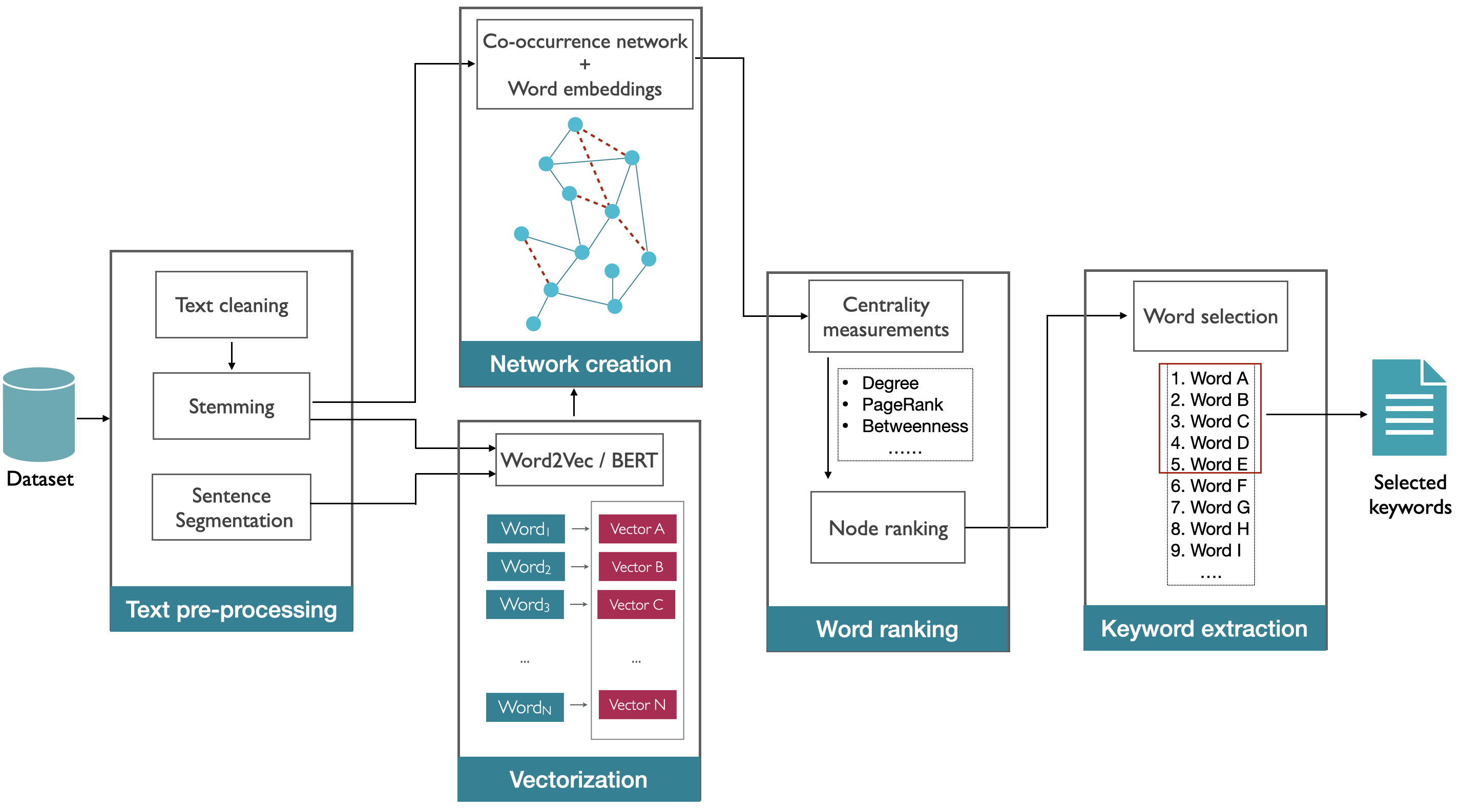}
    \caption{Architecture of our system for graph-based keyword extraction. The first step consists of pre-processing the input texts. Next, we obtain the vector representation of the words from the pre-processed datasets. Then co-occurrence networks are constructed considering two edges types (co-occurrence and embedding (virtual) edges). Several centrality measurements are calculated to rank words. Finally, the top-ranked words are considered as keywords for each document.}
   \label{fig:architecture}
\end{figure}

The workflow we considered for keyword extraction is shown in Figure~\ref{fig:architecture}.

\subsection{Pre-processing steps} \label{sec:preprop}

We applied some pre-processing steps before texts are represented as networks. We first performed sentence segmentation. We defined a sentence as any text portion which is separated by a period, exclamation or question mark. This step is needed because BERT embedding model requires the input documents to be separated into sentences. Next, we removed stopwords and punctuation marks. 
We finally applied text stemming to the remaining words so that words are converted into their singular, infinitive form.  This is important
to map related words into the same node. 
We did not consider text lemmatization because reference keywords from datasets were in their stemmed form.

\subsection{Word vectorization using word embeddings}\label{sec:methodology:embeddings}

Word embeddings models are a set of methods to represent words as dense vector representations. The idea behind these models is that words having similar meaning should have similar vector representations~\cite{mikolov2013efficient}. Word embeddings have been successfully used for several text applications such as information retrieval, question answering, document summarization and text classification~\cite{jiang2020combining, birunda2021review, shen2017word}. Here we use embeddings to establish links between similar nodes that do not co-occur in the text. The adopted method for including embedding edges as well as the construction process of the networks is described in Section~\ref{sec:methodology:network}.

There is a myriad of approaches to representing words as vectors. Methods to create word vectors include approaches based on neural networks, dimension reduction and probabilistic theory~\cite{almeida2019word}. Here we employed the following methods: 

\begin{itemize}

    \item \emph{Word2Vec}: This method is one of the first models to represent words as vectors~\cite{mikolov2013efficient}. Given a corpus, Word2Vec analyzes the words of each sentence and tries to predict neighbor words. For example, in the sentence ``The early bird catches the $X$'', Word2Vec can predict that the next word $X$ is ``worm'', based on the previous context. 
    This model uses a neural network with a single hidden layer. The neural network is trained with the documents of the corpus, then, for a given word $\alpha$, it is calculated the probability that each word of the vocabulary is a neighbor of $\alpha$. 
    Once the network is trained, the model uses the weights of the hidden layer. as word vectors. 
    Before the training stage, we defined different dimensions ($d$) for the word vectors. We generated vectors with $100 \leq d \leq 1,000$.
    %
    %
    Despite its simplicity and efficiency for various applications, Word2Vec has a significant weakness: it generates a unique vector for each word, regardless of word meaning and context, and this can generate noisy vectors, especially when representing ambiguous words. For example, the word ``apple" will have the same associated vector regardless of whether it refers to the apple fruit or the Apple technology company.  The BERT model addresses this problem as it generates different vectors for each word by taking into account the context in which the word appears.
    
    \item \emph{Bidirectional Encoder Representations from Transformer (BERT)}: This model creates representations using the context appearing before and after the target word. Then, once previously trained, it can be fine-tuned for several specific tasks~\cite{devlin2018bert}. BERT uses a multilayer model of transformers (self-attention modules), and  these structures allow learning attention weights of each word appearing before and after the target word. The model is pre-trained in two unsupervised tasks. In the first task, the model hides a percentage of the input tokens (words), and then it learns how to predict them. In the second task, the model selects two sentences, and then it predicts whether they are consecutive or not. Once the model has been pre-trained, it can be adjusted in a different task via a fine-tuning of parameters. We used this model to get the word vector representations of the keyword extraction datasets. We used the pre-trained model of BERT, which was previously trained over millions of texts. 
    For each sentence, we then obtained the representative vectors of the words composing that sentence. In this sense, each occurrence of the same word is represented by a different vector. The context of each occurrence is used to generate the vectors. 

\end{itemize}


{Recently, a large number of word embedding algorithms have been proposed to mathematically represent words and text segments. In this work, we used Word2Vec for being one of the first word embeddings models that were proposed as an improvement to the traditional vector space models. Furthermore, Word2Vec is a simple model whose training stage is fast compared to other techniques. Word2Vec has also been used quite successfully for small and large datasets. BERT is one of the first models to offer significant gain in performance compared to several models based on Word2Vec. Such gain lies in the fact that BERT and related models are capable of producing various vector representations of a word according to its context. In this sense, BERT is able to capture the polysemy of a word, which typically results in more accurate feature representations~\cite{devlin2018bert}.}


\subsection{Network construction}\label{sec:methodology:network}

\begin{figure}[h]
    \centering
    \includegraphics[width=1.0\textwidth]{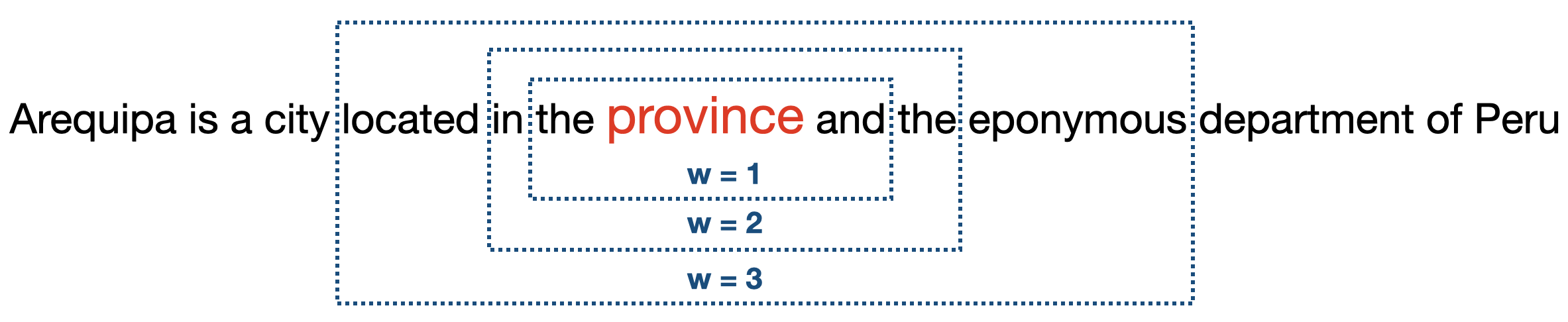}
    \caption{Example of how to find neighbors of a word for creating co-occurrence edges. In the sentence extracted from Wikipedia, we defined the neighbors of ``province'' according to a predefined window. If $w=1$, the immediate left and right side words (``the'' and ``and'') are considered. The window length $w=3$ includes the three words at the left and right side of the reference word ``province'': ``located'', ``in'', ``the'', ``and'', ``the'', and ``eponymous''.
    }
   \label{fig:text_example}
\end{figure}

After the pre-processing steps are applied, a graph representation is created. The motivation for representing text as complex networks is the simple, yet competitive and \emph{interpretable} results obtained in related text analysis tasks~\cite{cong2014approaching,cremades2022disentangling,akimushkin2018role,ferraz2018representation}. 
The adopted graph represents each word as a node. 
For the creation of the edges between two words, we defined two edges types: edges based on the neighborhood relationship of two words (co-occurrence edges), and edges based on the semantic similarity of the words (embedding edges). Differently from previous approaches, here we establish long-range edges that can not be obtained from adjacency relationships alone. This approach is a way to link words that are semantically related but do not share the same stem.  For co-occurrence edges, the following procedure was applied: we first defined a window value of size $w$. Edges linking two nodes are established for all words coexisting within the window. To build all edges, the window slides along the document.  Figure~\ref{fig:text_example} shows an example of edge formation. 
Here we considered $w=\{1,2,3\}$. Larger values of $w$ were not considered in order to avoid a large complexity in the computation of network measurements. We also did not observe, in preliminary experiments, a significant performance gain when considering larger contexts.

After the construction of the networks considering the co-occurrence relationships, the next step is the addition of  edges established via \emph{embeddings similarity}. This type of edge will also be referred to as \emph{virtual edges}. Let $E_t$ be the number of traditional co-occurrence edges. The number of virtual  edges included is $E_v = P E_t$. Here we considered a small percentage $P$ of additional edges, with $0 \leq P \leq 1$. The included \emph{virtual} edges are the $E_V$ most similar ones, according to the cosine similarity index. This strategy of enriching complex networks has been useful to provide more information in related applications~\cite{santos-etal-2017-enriching,Quispe2021}. This is particularly useful in short texts~\cite{santos-etal-2017-enriching}. We did not include more edges to avoid the complexity of analyzing denser networks. In addition, we did not find significant improvement in performance when the network is strongly connected.

\subsection{Network characterization} \label{sec:methodology:ranking}

The final step consists in using centrality measurements to rank the words according to the topological significance of the nodes from the network. Therefore, the best-ranked words (nodes) are chosen to be part of the resulting keyword list of the document. 
Centrality measurements are used to identify the most relevant nodes in a network. They are structural (or dynamical) attributes that indicate how central is a node according to a specific criterion. The identification of central nodes has been successfully used for various text applications. For example, \cite{Tohalino2018} used several traditional network measurements to identify the most important sentences in a sentence network. \cite{vega2019multi} modeled documents as word co-occurrence networks and used several centrality measurements to rank the words for the keyword extraction task. The network measurements were also used as features for classification and authorship attribution tasks. For example, \cite{Quispe2021} represented literary books as word co-occurrence networks and the centrality measurements of the most frequent words were considered as feature vectors. Then the selected vectors were used in a machine learning algorithm for authorship identification. Here, we evaluated traditional network measurements and their weighted versions. We also considered the accessibility metric owing to its relative success in text analysis~\cite{Tohalino2018}. Apart from the degree, we refer to the weighted version of metric $X$ as $X^{(w)}$.

\begin{enumerate}
 
    \item \emph{Degree} ($k$) and \emph{strength} ($s$): The node degree of a node is the number of edges that are connected to that node. In the case of weighted networks, the strength represents the sum of the weights of all the edges that are connected to the reference node.    
     
    \item \emph{PageRank} ($\pi$): This measurement considers a node $i$ as relevant if it is connected to other relevant nodes. The PageRank can be computed in a recursive way:
    \begin{equation}
        \label{eq:pr}
        \pi_i = \gamma \sum_j a_{ij} \frac{\pi_j}{k_j} + \beta, 
    \end{equation}
    where $\gamma$ and $\beta$ are used as damping factors, with $0 \leq \gamma \leq 1$ and $0 \leq \beta \leq 1$~\cite{langville2011google}. 
    {We also used a variation of this measurement that is based on the eigenvector centrality \emph{Eigenvector centrality} ($EV$).} 
    
    \item \emph{Betweenness} ($B$):  This metric is computed as the portion of shortest paths between two nodes that pass through a reference node. The betweenness centrality quantifies the relevance of a node to disseminate information~\cite{brandes2001faster}. It can also be used to identify words that are relevant even when they are not frequent~\cite{amancio2011comparing}. 
     
    \item \emph{Closeness} ($C$): This measurement tries to detect the nodes that can efficiently spread information through a network. It is defined as $C_i = N \sum_j 1/d_{ij}$, where $d_{ij}$ is the distance between $i$ and $j$, and $N$ is the number of nodes in the network. Nodes having high closeness value will have the shortest distances to all other nodes~\cite{sabidussi1966centrality}. Distance-based measurements have also been used to analyze texts~\cite{amancio2011comparing}. 
     
    \item \emph{Accessibility} ($A^{(h)}$): The accessibility metric quantifies the number of accessible nodes from an initial node using self-avoiding random walks of length $h$~\cite{Travencolo}. Nodes having a high accessibility also have effective access to more neighbors. This metric considers both the number of nodes at a given distance and the transition probabilities between the source and neighbor nodes. The accessibility can be evaluated considering different hierarchy levels. The levels can be set by specifying the length $h$ of the random walks~\cite{Travencolo}. To compute this metric for a reference node $i$, we first define $p^{(h)}(i,j)$ to denote the likelihood of reaching a node $j$ from an initial node $i$ in a self-avoiding random walk of length $h$. Then, the accessibility of $i$ is defined as the exponential of the true diversity of $p^{(h)}(i,j)$:  
    \begin{equation} \label{eq:accs}
        A_i^{(h)} = \exp \bigg(-\sum_j p^{(h)}(i,j) \log p^{(h)}(i,j) \bigg).
    \end{equation}
    This measurement has been used in several contexts to analyze texts, including in stylometric and semantic tasks~\cite{silva2022accessibility}. 
     
\end{enumerate}
 
We used each centrality measurement to assign different importance values for each word. Then,  the centrality values were used to rank the words. Therefore, the adopted methodology generated various word rankings according to the chosen network metrics. In Section~\ref{sec:results}, we reported the performance obtained for each network metric. After the word ranking step is performed, we selected the $N$ best-ranked words, where $N$ is the number of reference keywords. 
 
\subsection{Dataset}

We used publicly available datasets including the source texts and their gold-standard keywords defined by experts. The following datasets were chosen for their variability in size and sources. 
The \emph{Hult-2003} contains  title, keywords, and abstracts from scientific papers published between 1998 and 2002~\cite{hulth2003improved}. The documents were extracted from the \emph{Inspect Database of Physics and Engineering} papers~\cite{hulth2003improved}. This dataset contains $500$ abstracts as well as the set of keywords that were assigned by human annotators. The average size of the documents from this dataset is about $123$ words. The \emph{Marujo-2012} dataset comprises $450$ web news stories on subjects such as business, culture, sport, and technology~\cite{Marujo2012399}. 
The mean document size is 452 words. Finally, we also used the Semeval-2010~\cite{kim2010semeval}. This dataset comprises scientific papers that were extracted from the ACM Digital Library. We considered the full content of $100$ papers and their corresponding keywords assigned by both authors and readers~\cite{kim2010semeval}. The average document length is $8,168$ words.  In Table~\ref{tab:datasets} we provide a summary indicating the main attributes of each dataset.

\begin{table}[h]
\caption{Statistical information from datasets for the keyword extraction task. $|D|$ represents the number of documents. We also show the average number of tokens ($\langle W \rangle$), sentences $\langle S \rangle$) and vocabulary size ($\langle U \rangle$). $\langle K \rangle$ is the average number of reference keywords assigned per document.}
\begin{tabular}{c|c|c|c|c|c|c}
\hline \hline
\textbf{Dataset} & \textbf{Description} & \textbf{$|D|$} & \textbf{$W_{avg}$} & \textbf{$U_{avg}$} & \textbf{$S_{avg}$} & \textbf{$K_{avg}$}  \\ \hline
Hult-2003 & Paper abstracts & $500$ & $123.12$ & $73.25$ & $5.14$ & $18.83$   \\ 
Marujo-2012 & Web news stories & $450$ & $452.36$ & $223.33$ & $20.74$ & $52.79$  \\ 
SemEval-2010 & Full papers & $100$ & $8168.49$ & $1387.47$ & $393.80$ & $23.34$   \\ \hline \hline
\end{tabular}
\label{tab:datasets}
\end{table}

\section{Results and discussion}\label{sec:results}

In this section, we analyze whether our hypothesis that the inclusion of virtual edges can improve the performance of co-occurrence networks in detecting keywords. In Section~\ref{sec:results:measures}, an analysis of the effect of parameter variation on the performance is provided. In Sections~\ref{sec:results:w2v} and \ref{sec:results:bert}, we detail the results obtained with Word2Vec and BERT, respectively. Finally, we show in Section~\ref{sec:results:summary} a summary of the obtained results.

\subsection{Parameter analysis} \label{sec:results:measures}

In this section, we investigate whether the proposed extension of traditional word adjacency networks can lead to optimized results. In this section, we analyze if the performance is improved when we vary the model parameters. We are particularly interested in the performance analysis when varying both the window length ($w$)  and the number of virtual edges ($P$). We focus our analysis on the results obtained for the Word2vec model, since similar results have been found with BERT (see next sections). 

\begin{figure}[h]
    \centering
    \includegraphics[width=1.0\textwidth]{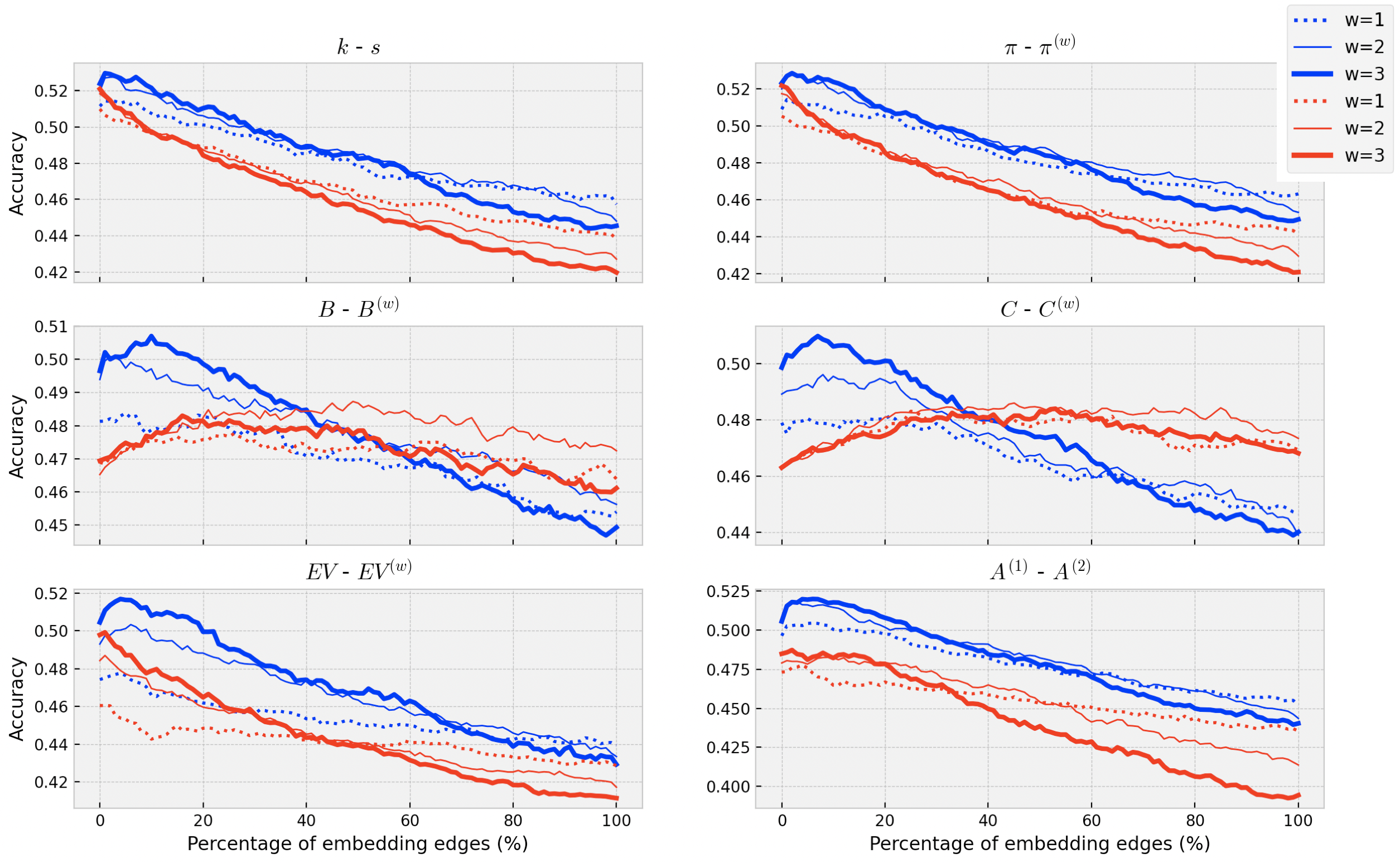}
    \caption{Comparison of the performance of each centrality measurement based on the Word2Vec model for the \emph{Hult-2003} dataset. For all subplots except the Accessibility metric, the blue lines represent unweighted measurements, while the red lines are weighted measurements. In the case of the Accessibility centrality, the blue lines describe the $A^{(1)}$ metric and the red lines represent the $A^{(2)}$ metric. We also evaluated the window length ($w = \{1, 2, 3\} $) for the network creation step: dotted lines are used when the value $w = 1$ is established, while thicker lines represent values for larger values of $w$.}
  \label{fig:hult_results}
\end{figure}

In Figure~\ref{fig:hult_results} we show, for the Hult-2003 dataset, the performance obtained for different network metrics. We considered distinct model parameters, with the window length being represented by different curves and $P$ represented on the x-axis. The effect of considering edge weights was also considered. Note that the traditional word adjacency (unweighted) model is represented by dotted blue curves. The results observed in this dataset reveal that the best results are achieved with the largest window ($w=3$) for all of the considered metrics. This means that a wider context does provide a better model for detecting keywords. Concerning the comparison of weighted and unweighted metrics, the best result considering all parameter combinations is always achieved with the unweighted version of the metrics. Most importantly, we also see that, considering the largest context and the unweighted version ($w=3$), the inclusion of additional edges is also able to improve the performance of the methods.  In all considered unweighted measurements, the inclusion of a few virtual edges can lead to optimized results. Interestingly, one should observe though that the inclusion of a large number of edges can cause a loss of relevant information. Whenever $P > 0.60$, the performance tends to decrease. The results observed in the figures also show that the best accuracy rates occurs typically for $0 \leq P \leq 0.20$.

\begin{figure}[h]
    \centering
    \includegraphics[width=1.0\textwidth]{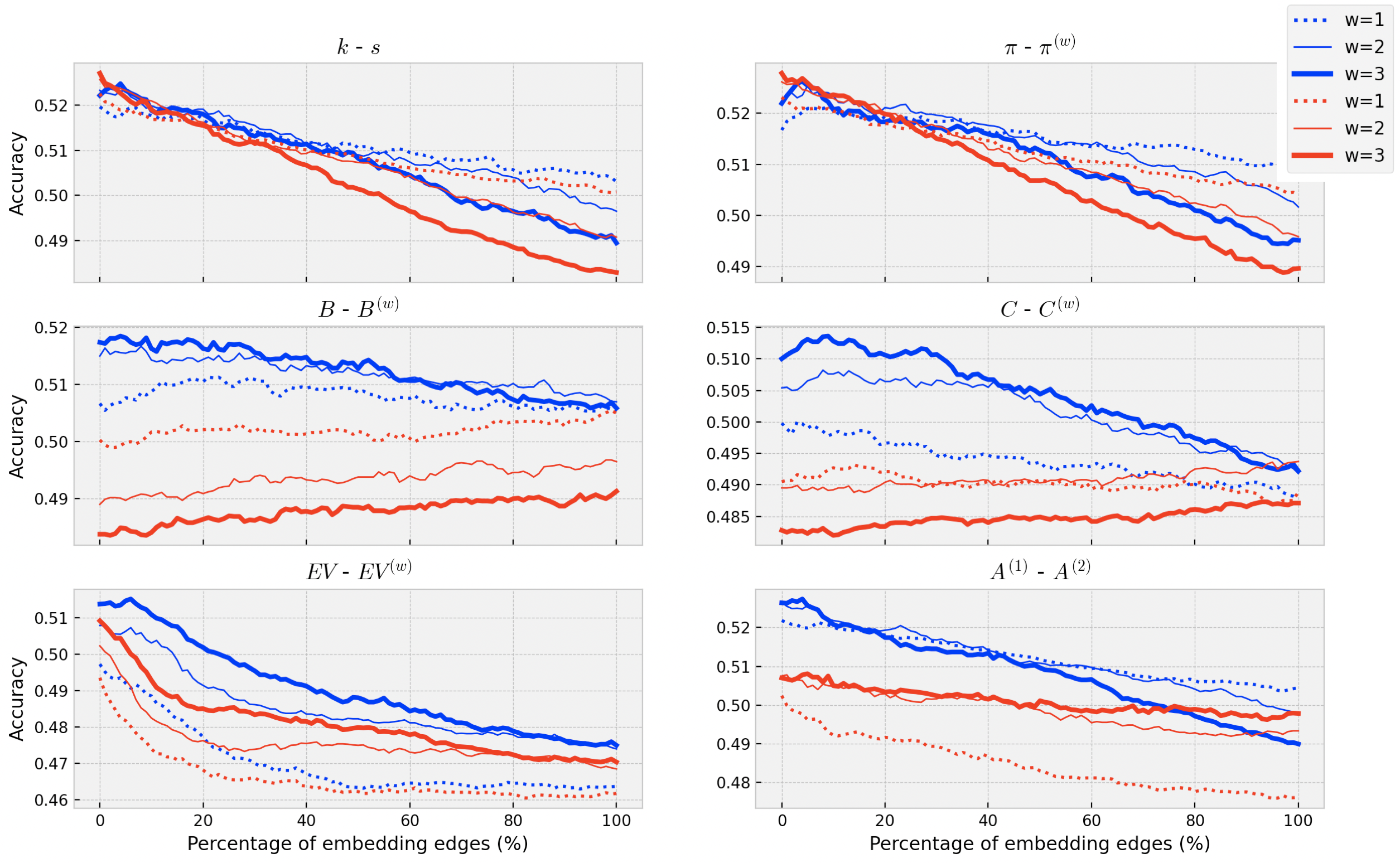}
    \caption{Comparison of the performance of each centrality measurement based on the Word2Vec model for the \emph{Marujo-2012} dataset. For all subplots except the Accessibility metric, the blue lines represent unweighted measurements, while the red lines stand for measurements based on weights. In the case of the Accessibility centrality, the blue lines describe the $A^{(1)}$ metric and the red lines represent the $A^{(2)}$ metric. We also evaluated the window length ($w = \{1, 2, 3\} $) for the network creation step: dotted lines are used when the value $w = 1$ is established, while thicker lines represent values for $w$ larger than 1.}
  \label{fig:marujo_results}
\end{figure}

When analyzing the Marujo-2012 dataset (see Figure \ref{fig:marujo_results}), some differences can be observed. While the inclusion of virtual edges can improve the results of some models, one observes that the best results are obtained with the largest window length. Conversely, the role of including additional edges depends on the considered model. For both degree and PageRank, the best results were found with the weighted version and the largest window ($w=3$) (see the red curve). In this case, the inclusion of additional edges hampers the performance. For all other metrics, the best results were found with the largest window length and the unweighted version when a small percentage of edges is included. 

\begin{figure}[h]
    \centering
    \includegraphics[width=1.0\textwidth]{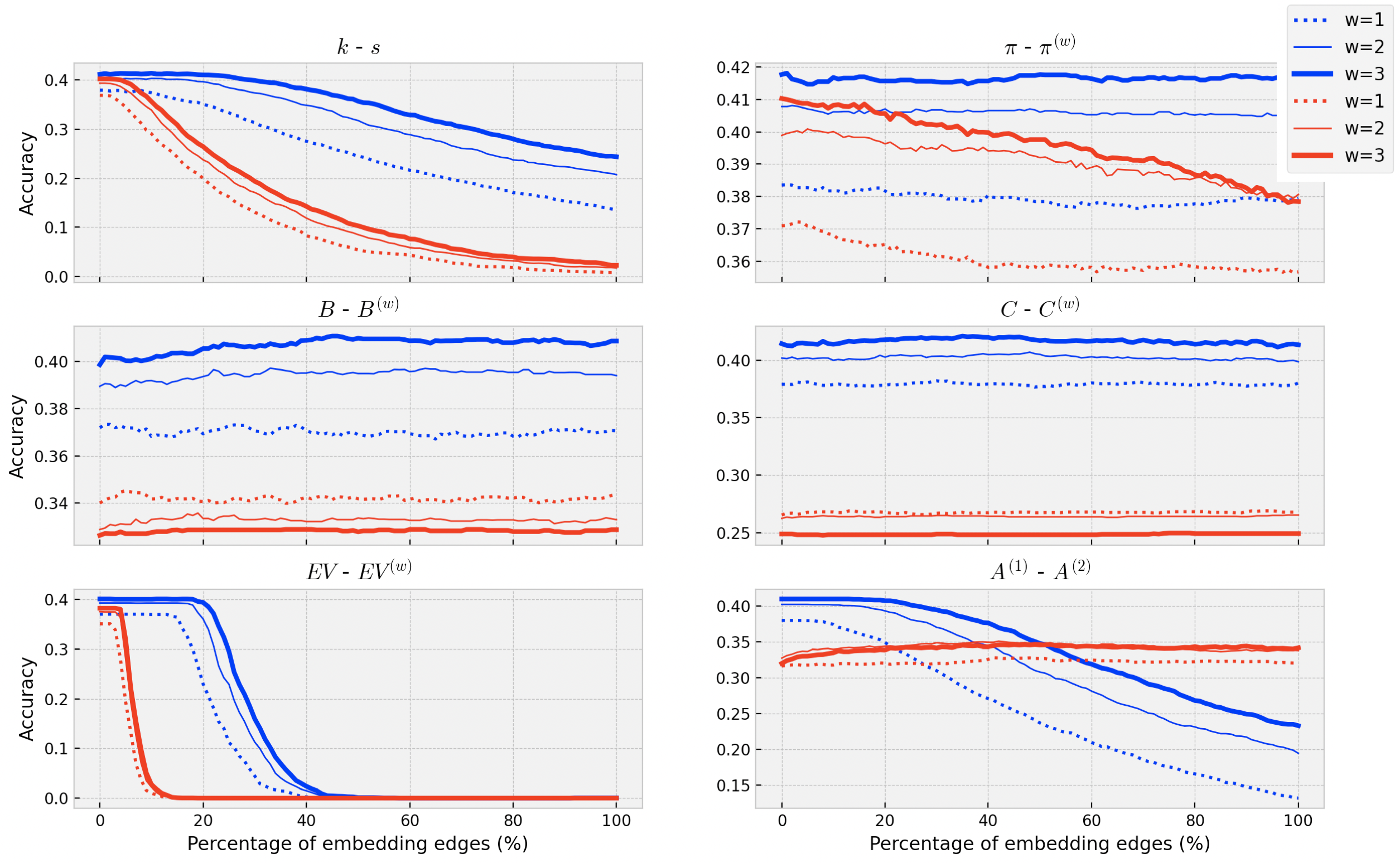}
    \caption{Comparison of the performance of each centrality measurement based on the Word2Vec model for the \emph{SemEval-2010 dataset}. For all subplots except the Accessibility metric, the blue lines represent unweighted measurements, while the red lines stand for measurements based on weights. In the case of the Accessibility centrality, the blue lines describe the $A^{(1)}$ metric and the red lines represent the $A^{(2)}$ metric. We also evaluated the use of windows ($w = \{1, 2, 3\} $) for the network creation step: dotted lines are used when the value $w=1$ is established, while thicker lines represent values for $w$ larger than 1. }
  \label{fig:semeval_results}
\end{figure}

When one observes the results for the SemEval-2010 dataset in Figure \ref{fig:semeval_results}, all best results were found with the unweighted version of the model considering $w=3$. However, differently from the other datasets, for almost all metrics the inclusion of virtual edges does not improve the performance of the keyword detection. The performance with PageRank and closeness are not \emph{positively} affected by the inclusion of virtual edges, when analyzing the blue curves with the highest performance. 
The degree, eigenvector centrality and accessibility are negatively affected if several virtual edges are included. Surprisingly, the informativeness of the model even disappears when more than 50\% of virtual edges are included for the eigenvector centrality. The betweenness centrality seems to be the only metric being improved -- marginally -- when virtual edges are included.

 
All in all the results show that the parameter behavior seems to depend on the considered dataset. In short texts (Hult-2003), the importance of including virtual edges is clearly observed. This happens because when short texts are modeled as co-occurrence networks, the generated line is almost a graph line. As a consequence, the topological information is not able to detect keywords, since all concepts will have the same topological information. In this case, the use of virtual edges is essential to identify the hidden information in short texts. Therefore, the results suggest that the proposed methodology can be useful to analyze short texts. 
Despite the above differences, the optimized results are almost always obtained when using a large window length ($w=3$). The weighted metrics did not provide a significant gain in performance over their unweighted versions.
 
\subsection{Performance analysis using the Word2Vec model}\label{sec:results:w2v}

%
Table~\ref{tab:w2v_results} depicts the results of the evaluation of the Word2Vec model considering $100$, $300$ and $500$  dimensions (result not shown). We did not include the results obtained with larger dimensions because the observed performance decreases compared to smaller dimensions. We also show the performance of each vector size when the window parameter ($w$) was considered. For each measurement, we also show the percentage of embedding edge insertion that yielded the highest accuracy rates ($P$) and the highest accuracy observed with the proposed model (Acc.). We defined two additional quantities $\Gamma_1$ and $\Gamma_2$, which are defined as
\begin{equation}
    \Gamma_1 = \frac{\textrm{Acc} - \textrm{Acc}^{\textrm{(tr)}}}{\textrm{Acc}^{\textrm{(tr)}}}, 
\end{equation}
\begin{equation}
    \Gamma_2 = \frac{\textrm{Acc} - \textrm{Acc}^{\textrm{(w)}}}{\textrm{Acc}^{\textrm{(w)}}}. 
\end{equation}
$\textrm{Acc}^{\textrm{(tr)}}$ corresponds to the accuracy obtained with the traditional co-occurrence model~\cite{sulis2022exploiting} (i.e, our model with $P=0$ and $w=1$). $\textrm{Acc}^{\textrm{(tr)}}$ corresponds to accuracy obtained with the model considering only co-occurrence links~\cite{vega2019multi} (i.e., our model with $P=0$). Thus, $\Gamma_1$ and $\Gamma_2$ quantify the gain in performance when important features of the model are disregarded.  

\begin{table}[]
\caption{Performance based on the Word2Vec model for edge embedding creation. Acc. represents the highest accuracy rate for the considered set of parameters. For this analysis, we experimented with different dimensions ($d$) of the embedding vectors.}
\label{tab:w2v_results}
{\footnotesize
\begin{tabular}{cc|ccccc|ccccc|ccccc|}
\cline{3-17} & & \multicolumn{5}{c|}{\textbf{d = 100}} & \multicolumn{5}{c|}{\textbf{d = 300}} & \multicolumn{5}{c|}{\textbf{d = 500}} \\ \hline

\multicolumn{1}{|c|}{\textbf{Dataset}} & \textbf{Meas.} & \multicolumn{1}{c}{\textbf{P}} & \multicolumn{1}{c}{\textbf{w}} & \multicolumn{1}{c}{\textbf{$\Gamma_1$}} & \multicolumn{1}{c}{\textbf{$\Gamma_2$}} & \textbf{Acc.} & \multicolumn{1}{c}{\textbf{P}} & \multicolumn{1}{c}{\textbf{w}} & \multicolumn{1}{c}{\textbf{$\Gamma_1$}} & \multicolumn{1}{c}{\textbf{$\Gamma_2$}} & \textbf{Acc.} & \multicolumn{1}{c}{\textbf{P}} & \multicolumn{1}{c}{\textbf{w}} & \multicolumn{1}{c}{\textbf{$\Gamma_1$}} & \multicolumn{1}{c}{\textbf{$\Gamma_2$}} & \textbf{Acc.} \\ \hline

\multicolumn{1}{|c|}{\multirow{12}{*}{Hult-2003}} & $k$ & \multicolumn{1}{c}{2} & \multicolumn{1}{c}{3} & \multicolumn{1}{c}{0.03} & \multicolumn{1}{c}{0.01} & 0.5280 & \multicolumn{1}{c}{1} & \multicolumn{1}{c}{3} & \multicolumn{1}{c}{0.04} & \multicolumn{1}{c}{0.01} & 0.5296 & \multicolumn{1}{c}{2} & \multicolumn{1}{c}{3} & \multicolumn{1}{c}{0.04} & \multicolumn{1}{c}{0.02} & 0.5288\\

\multicolumn{1}{|c|}{} & $s$ & \multicolumn{1}{c}{0} & \multicolumn{1}{c}{3} & \multicolumn{1}{c}{0.03} &  \multicolumn{1}{c}{--} & 0.5184 & \multicolumn{1}{c}{0} & \multicolumn{1}{c}{3} & \multicolumn{1}{c}{0.02} & \multicolumn{1}{c}{--} & 0.5209 & \multicolumn{1}{c}{0} & \multicolumn{1}{c}{3} & \multicolumn{1}{c}{0.03} & \multicolumn{1}{c}{--} & 0.5223 \\ 

\multicolumn{1}{|c|}{} & $\pi$ & \multicolumn{1}{c}{2} & \multicolumn{1}{c}{3} & \multicolumn{1}{c}{0.04} & \multicolumn{1}{c}{0.01} & 0.5274 & \multicolumn{1}{c}{2} & \multicolumn{1}{c}{3} & \multicolumn{1}{c}{0.04} & \multicolumn{1}{c}{0.01} & 0.5283 & \multicolumn{1}{c}{2} & \multicolumn{1}{c}{3} & \multicolumn{1}{c}{0.04} & \multicolumn{1}{c}{0.01} & 0.5282 \\ 

\multicolumn{1}{|c|}{} & $\pi^{(w)}$ & \multicolumn{1}{c}{0} & \multicolumn{1}{c}{3} & \multicolumn{1}{c}{0.03} & \multicolumn{1}{c}{--} & 0.5178 & \multicolumn{1}{c}{0} & \multicolumn{1}{c}{3} & \multicolumn{1}{c}{0.03} & \multicolumn{1}{c}{--} & 0.5216 & \multicolumn{1}{c}{0} & \multicolumn{1}{c}{3} & \multicolumn{1}{c}{0.04} & \multicolumn{1}{c}{--} & 0.5237 \\ 

\multicolumn{1}{|c|}{} & $B$ & \multicolumn{1}{c}{8} & \multicolumn{1}{c}{3} & \multicolumn{1}{c}{0.05} & \multicolumn{1}{c}{0.01} & 0.5030 & \multicolumn{1}{c}{10} & \multicolumn{1}{c}{3} & \multicolumn{1}{c}{0.05} & \multicolumn{1}{c}{0.02} & 0.5070 & \multicolumn{1}{c}{7} & \multicolumn{1}{c}{3} & \multicolumn{1}{c}{0.04} & \multicolumn{1}{c}{0.01} & 0.5036 \\ 

\multicolumn{1}{|c|}{} & $B^{(w)}$ & \multicolumn{1}{c}{22} & \multicolumn{1}{c}{2} & \multicolumn{1}{c}{0.03} & \multicolumn{1}{c}{0.03} & 0.4876 & \multicolumn{1}{c}{49} & \multicolumn{1}{c}{2} & \multicolumn{1}{c}{0.04} & \multicolumn{1}{c}{0.04} & 0.4873 & \multicolumn{1}{c}{25} & \multicolumn{1}{c}{2} & \multicolumn{1}{c}{0.04} & \multicolumn{1}{c}{0.04} & 0.4882 \\ 

\multicolumn{1}{|c|}{} & $C$ & \multicolumn{1}{c}{3} & \multicolumn{1}{c}{3} & \multicolumn{1}{c}{0.06} & \multicolumn{1}{c}{0.01} & 0.5063 & \multicolumn{1}{c}{7} & \multicolumn{1}{c}{3} & \multicolumn{1}{c}{0.06} & \multicolumn{1}{c}{0.02} & 0.5098 & \multicolumn{1}{c}{5} & \multicolumn{1}{c}{3} & \multicolumn{1}{c}{0.06} & \multicolumn{1}{c}{0.02} & 0.5094 \\ 

\multicolumn{1}{|c|}{} & $C^{(w)}$ & \multicolumn{1}{c}{26} & \multicolumn{1}{c}{2} & \multicolumn{1}{c}{0.04} & \multicolumn{1}{c}{0.04} & 0.4879 & \multicolumn{1}{c}{45} & \multicolumn{1}{c}{2} & \multicolumn{1}{c}{0.05} & \multicolumn{1}{c}{0.05} & 0.4860 & \multicolumn{1}{c}{31} & \multicolumn{1}{c}{2} & \multicolumn{1}{c}{0.05} & \multicolumn{1}{c}{0.05} & 0.4875 \\ 

\multicolumn{1}{|c|}{} & $EV$ & \multicolumn{1}{c}{6} & \multicolumn{1}{c}{3} & \multicolumn{1}{c}{0.09} & \multicolumn{1}{c}{0.02} & 0.5173 & \multicolumn{1}{c}{4} & \multicolumn{1}{c}{3} & \multicolumn{1}{c}{0.09} & \multicolumn{1}{c}{0.02} & 0.5169 & \multicolumn{1}{c}{5} & \multicolumn{1}{c}{3} & \multicolumn{1}{c}{0.09} & \multicolumn{1}{c}{0.02} & 0.5154 \\ 

\multicolumn{1}{|c|}{} & $EV^{(w)}$ & \multicolumn{1}{c}{1} & \multicolumn{1}{c}{3} & \multicolumn{1}{c}{0.08} & \multicolumn{1}{c}{--} & 0.4999 & \multicolumn{1}{c}{1} & \multicolumn{1}{c}{3} & \multicolumn{1}{c}{0.08} & \multicolumn{1}{c}{--} & 0.4992 & \multicolumn{1}{c}{0} & \multicolumn{1}{c}{3} & \multicolumn{1}{c}{0.09} & \multicolumn{1}{c}{--} & 0.4995 \\ 

\multicolumn{1}{|c|}{} & $A^{(1)}$ & \multicolumn{1}{c}{5} & \multicolumn{1}{c}{3} & \multicolumn{1}{c}{0.05} & \multicolumn{1}{c}{0.02} & 0.5199 & \multicolumn{1}{c}{6} & \multicolumn{1}{c}{3} & \multicolumn{1}{c}{0.05} & \multicolumn{1}{c}{0.03} & 0.5200 & \multicolumn{1}{c}{6} & \multicolumn{1}{c}{3} & \multicolumn{1}{c}{0.06} & \multicolumn{1}{c}{0.02} & 0.5198 \\ 

\multicolumn{1}{|c|}{} & $A^{(2)}$ & \multicolumn{1}{c}{2} & \multicolumn{1}{c}{3} & \multicolumn{1}{c}{0.04} & \multicolumn{1}{c}{0.01} & 0.4880 & \multicolumn{1}{c}{2} & \multicolumn{1}{c}{3} & \multicolumn{1}{c}{0.03} & \multicolumn{1}{c}{--} & 0.4872 & \multicolumn{1}{c}{3} & \multicolumn{1}{c}{3} & \multicolumn{1}{c}{0.03} & \multicolumn{1}{c}{0.01} & 0.4866 \\ \hline

\multicolumn{1}{|c|}{\multirow{12}{*}{Marujo-2012}} & $k$ & \multicolumn{1}{c}{2} & \multicolumn{1}{c}{3} & \multicolumn{1}{c}{0.01} & \multicolumn{1}{c}{0.01} & 0.5275 & \multicolumn{1}{c}{4} & \multicolumn{1}{c}{3} & \multicolumn{1}{c}{0.01} & \multicolumn{1}{c}{--} & 0.5247 & \multicolumn{1}{c}{2} & \multicolumn{1}{c}{3} & \multicolumn{1}{c}{0.01} & \multicolumn{1}{c}{--} & 0.5278\\

\multicolumn{1}{|c|}{} & $s$ & \multicolumn{1}{c}{0} & \multicolumn{1}{c}{3} & \multicolumn{1}{c}{0.01} & \multicolumn{1}{c}{--} & 0.5279 & \multicolumn{1}{c}{0} & \multicolumn{1}{c}{3} & \multicolumn{1}{c}{0.01} & \multicolumn{1}{c}{--} & 0.5270 & \multicolumn{1}{c}{0} & \multicolumn{1}{c}{3} & \multicolumn{1}{c}{0.01} & \multicolumn{1}{c}{--} & 0.5276 \\ 

\multicolumn{1}{|c|}{} & $\pi$ & \multicolumn{1}{c}{4} & \multicolumn{1}{c}{3} & \multicolumn{1}{c}{0.02} & \multicolumn{1}{c}{0.01} & 0.5258 & \multicolumn{1}{c}{4} & \multicolumn{1}{c}{3} & \multicolumn{1}{c}{0.02} & \multicolumn{1}{c}{0.01} & 0.5262 & \multicolumn{1}{c}{5} & \multicolumn{1}{c}{3} & \multicolumn{1}{c}{0.02} & \multicolumn{1}{c}{0.01} & 0.5266 \\ 

\multicolumn{1}{|c|}{} & $\pi^{(w)}$ & \multicolumn{1}{c}{0} & \multicolumn{1}{c}{3} & \multicolumn{1}{c}{0.01} & \multicolumn{1}{c}{--} & 0.5289 & \multicolumn{1}{c}{0} & \multicolumn{1}{c}{3} & \multicolumn{1}{c}{0.01} & \multicolumn{1}{c}{--} & 0.5278 & \multicolumn{1}{c}{0} & \multicolumn{1}{c}{3} & \multicolumn{1}{c}{0.01} & \multicolumn{1}{c}{--} & 0.5286 \\ 

\multicolumn{1}{|c|}{} & $B$ & \multicolumn{1}{c}{5} & \multicolumn{1}{c}{3} & \multicolumn{1}{c}{0.03} & \multicolumn{1}{c}{--} & 0.5195 & \multicolumn{1}{c}{4} & \multicolumn{1}{c}{3} & \multicolumn{1}{c}{0.02} & \multicolumn{1}{c}{--} & 0.5185 & \multicolumn{1}{c}{4} & \multicolumn{1}{c}{3} & \multicolumn{1}{c}{0.03} & \multicolumn{1}{c}{--} & 0.5193 \\ 

\multicolumn{1}{|c|}{} & $B^{(w)}$ & \multicolumn{1}{c}{43} & \multicolumn{1}{c}{1} & \multicolumn{1}{c}{0.01} & \multicolumn{1}{c}{0.01} & 0.5029 & \multicolumn{1}{c}{99} & \multicolumn{1}{c}{1} & \multicolumn{1}{c}{0.01} & \multicolumn{1}{c}{0.01} & 0.5055 & \multicolumn{1}{c}{31} & \multicolumn{1}{c}{1} & \multicolumn{1}{c}{0.01} & \multicolumn{1}{c}{0.01} & 0.5031 \\ 

\multicolumn{1}{|c|}{} & $C$ & \multicolumn{1}{c}{7} & \multicolumn{1}{c}{3} & \multicolumn{1}{c}{0.03} & \multicolumn{1}{c}{0.01} & 0.5143 & \multicolumn{1}{c}{9} & \multicolumn{1}{c}{3} & \multicolumn{1}{c}{0.03} & \multicolumn{1}{c}{0.01} & 0.5136 & \multicolumn{1}{c}{19} & \multicolumn{1}{c}{3} & \multicolumn{1}{c}{0.03} & \multicolumn{1}{c}{0.01} & 0.5141 \\ 

\multicolumn{1}{|c|}{} & $C^{(w)}$ & \multicolumn{1}{c}{87} & \multicolumn{1}{c}{2} & \multicolumn{1}{c}{--} & \multicolumn{1}{c}{--} & 0.4931 & \multicolumn{1}{c}{100} & \multicolumn{1}{c}{2} & \multicolumn{1}{c}{0.01} & \multicolumn{1}{c}{0.01} & 0.4937 & \multicolumn{1}{c}{77} & \multicolumn{1}{c}{2} & \multicolumn{1}{c}{--} & \multicolumn{1}{c}{--} & 0.4923 \\ 

\multicolumn{1}{|c|}{} & $EV$ & \multicolumn{1}{c}{7} & \multicolumn{1}{c}{3} & \multicolumn{1}{c}{0.04} & \multicolumn{1}{c}{--} & 0.5160 & \multicolumn{1}{c}{6} & \multicolumn{1}{c}{3} & \multicolumn{1}{c}{0.04} & \multicolumn{1}{c}{--} & 0.5151 & \multicolumn{1}{c}{6} & \multicolumn{1}{c}{3} & \multicolumn{1}{c}{0.04} & \multicolumn{1}{c}{--} & 0.5155 \\ 

\multicolumn{1}{|c|}{} & $EV^{(w)}$ & \multicolumn{1}{c}{0} & \multicolumn{1}{c}{3} & \multicolumn{1}{c}{0.03} & \multicolumn{1}{c}{--} & 0.5095 & \multicolumn{1}{c}{0} & \multicolumn{1}{c}{3} & \multicolumn{1}{c}{0.03} & \multicolumn{1}{c}{--} & 0.5091 & \multicolumn{1}{c}{0} & \multicolumn{1}{c}{3} & \multicolumn{1}{c}{0.03} & \multicolumn{1}{c}{--} & 0.5105 \\ 

\multicolumn{1}{|c|}{} & $A^{(1)}$ & \multicolumn{1}{c}{1} & \multicolumn{1}{c}{3} & \multicolumn{1}{c}{0.01} & \multicolumn{1}{c}{--} & 0.5275 & \multicolumn{1}{c}{4} & \multicolumn{1}{c}{3} & \multicolumn{1}{c}{0.01} & \multicolumn{1}{c}{--} & 0.5274 & \multicolumn{1}{c}{2} & \multicolumn{1}{c}{3} & \multicolumn{1}{c}{0.01} & \multicolumn{1}{c}{--} & 0.5275 \\ 

\multicolumn{1}{|c|}{} & $A^{(2)}$ & \multicolumn{1}{c}{5} & \multicolumn{1}{c}{3} & \multicolumn{1}{c}{0.01} & \multicolumn{1}{c}{--} & 0.5094 & \multicolumn{1}{c}{5} & \multicolumn{1}{c}{3} & \multicolumn{1}{c}{0.01} & \multicolumn{1}{c}{--} & 0.5081 & \multicolumn{1}{c}{3} & \multicolumn{1}{c}{3} & \multicolumn{1}{c}{0.01} & \multicolumn{1}{c}{--} & 0.5097 \\ \hline

\multicolumn{1}{|c|}{\multirow{12}{*}{SemEval-2010}} & $k$ & \multicolumn{1}{c}{0} & \multicolumn{1}{c}{3} & \multicolumn{1}{c}{0.09} & \multicolumn{1}{c}{--} & 0.4140 & \multicolumn{1}{c}{10} & \multicolumn{1}{c}{3} & \multicolumn{1}{c}{0.09} & \multicolumn{1}{c}{0.01} & 0.4140 & \multicolumn{1}{c}{2} & \multicolumn{1}{c}{3} & \multicolumn{1}{c}{0.10} & \multicolumn{1}{c}{--} & 0.4144\\

\multicolumn{1}{|c|}{} & $s$ & \multicolumn{1}{c}{0} & \multicolumn{1}{c}{3} & \multicolumn{1}{c}{0.09} & \multicolumn{1}{c}{--} & 0.4039 & \multicolumn{1}{c}{0} & \multicolumn{1}{c}{3} & \multicolumn{1}{c}{0.09} & \multicolumn{1}{c}{--} & 0.4024 & \multicolumn{1}{c}{0} & \multicolumn{1}{c}{3} & \multicolumn{1}{c}{0.09} & \multicolumn{1}{c}{--} & 0.4035 \\ 

\multicolumn{1}{|c|}{} & $\pi$ & \multicolumn{1}{c}{0} & \multicolumn{1}{c}{3} & \multicolumn{1}{c}{0.09} & \multicolumn{1}{c}{--} & 0.4177 & \multicolumn{1}{c}{1} & \multicolumn{1}{c}{3} & \multicolumn{1}{c}{0.09} & \multicolumn{1}{c}{--} & 0.4181 & \multicolumn{1}{c}{48} & \multicolumn{1}{c}{3} & \multicolumn{1}{c}{0.09} & \multicolumn{1}{c}{--} & 0.4185 \\ 

\multicolumn{1}{|c|}{} & $\pi^{(w)}$ & \multicolumn{1}{c}{3} & \multicolumn{1}{c}{3} & \multicolumn{1}{c}{0.10} & \multicolumn{1}{c}{--} & 0.4111 & \multicolumn{1}{c}{0} & \multicolumn{1}{c}{3} & \multicolumn{1}{c}{0.11} & \multicolumn{1}{c}{--} & 0.4103 & \multicolumn{1}{c}{1} & \multicolumn{1}{c}{3} & \multicolumn{1}{c}{0.10} & \multicolumn{1}{c}{--} & 0.4106 \\ 

\multicolumn{1}{|c|}{} & $B$ & \multicolumn{1}{c}{74} & \multicolumn{1}{c}{3} & \multicolumn{1}{c}{0.10} & \multicolumn{1}{c}{0.03} & 0.4100 & \multicolumn{1}{c}{45} & \multicolumn{1}{c}{3} & \multicolumn{1}{c}{0.10} & \multicolumn{1}{c}{0.03} & 0.4106 & \multicolumn{1}{c}{43} & \multicolumn{1}{c}{3} & \multicolumn{1}{c}{0.10} & \multicolumn{1}{c}{0.03} & 0.4104 \\ 

\multicolumn{1}{|c|}{} & $B^{(w)}$ & \multicolumn{1}{c}{26} & \multicolumn{1}{c}{1} & \multicolumn{1}{c}{0.01} & \multicolumn{1}{c}{0.01} & 0.3521 & \multicolumn{1}{c}{5} & \multicolumn{1}{c}{1} & \multicolumn{1}{c}{0.02} & \multicolumn{1}{c}{0.02} & 0.3452 & \multicolumn{1}{c}{23} & \multicolumn{1}{c}{1} & \multicolumn{1}{c}{0.01} & \multicolumn{1}{c}{0.01} & 0.3458 \\ 

\multicolumn{1}{|c|}{} & $C$ & \multicolumn{1}{c}{46} & \multicolumn{1}{c}{3} & \multicolumn{1}{c}{0.10} & \multicolumn{1}{c}{0.01} & 0.4179 & \multicolumn{1}{c}{35} & \multicolumn{1}{c}{3} & \multicolumn{1}{c}{0.11} & \multicolumn{1}{c}{0.02} & 0.4208 & \multicolumn{1}{c}{35} & \multicolumn{1}{c}{3} & \multicolumn{1}{c}{0.11} & \multicolumn{1}{c}{0.02} & 0.4200 \\ 

\multicolumn{1}{|c|}{} & $C^{(w)}$ & \multicolumn{1}{c}{64} & \multicolumn{1}{c}{1} & \multicolumn{1}{c}{0.02} & \multicolumn{1}{c}{0.02} & 0.2743 & \multicolumn{1}{c}{93} & \multicolumn{1}{c}{1} & \multicolumn{1}{c}{0.01} & \multicolumn{1}{c}{0.01} & 0.2690 & \multicolumn{1}{c}{50} & \multicolumn{1}{c}{1} & \multicolumn{1}{c}{0.01} & \multicolumn{1}{c}{0.01} & 0.2682 \\ 

\multicolumn{1}{|c|}{} & $EV$ & \multicolumn{1}{c}{13} & \multicolumn{1}{c}{3} & \multicolumn{1}{c}{0.08} & \multicolumn{1}{c}{--} & 0.4011 & \multicolumn{1}{c}{0} & \multicolumn{1}{c}{3} & \multicolumn{1}{c}{0.08} & \multicolumn{1}{c}{--} & 0.4007 & \multicolumn{1}{c}{0} & \multicolumn{1}{c}{3} & \multicolumn{1}{c}{0.08} & \multicolumn{1}{c}{--} & 0.4007 \\ 

\multicolumn{1}{|c|}{} & $EV^{(w)}$ & \multicolumn{1}{c}{0} & \multicolumn{1}{c}{3} & \multicolumn{1}{c}{0.08} & \multicolumn{1}{c}{--} & 0.3807 & \multicolumn{1}{c}{0} & \multicolumn{1}{c}{3} & \multicolumn{1}{c}{0.09} & \multicolumn{1}{c}{--} & 0.3820 & \multicolumn{1}{c}{0} & \multicolumn{1}{c}{3} & \multicolumn{1}{c}{0.09} & \multicolumn{1}{c}{--} & 0.3835 \\ 

\multicolumn{1}{|c|}{} & $A^{(1)}$ & \multicolumn{1}{c}{0} & \multicolumn{1}{c}{3} & \multicolumn{1}{c}{0.08} & \multicolumn{1}{c}{--} & 0.4088 & \multicolumn{1}{c}{0} & \multicolumn{1}{c}{3} & \multicolumn{1}{c}{0.08} & \multicolumn{1}{c}{--} & 0.4099 & \multicolumn{1}{c}{7} & \multicolumn{1}{c}{3} & \multicolumn{1}{c}{0.08} & \multicolumn{1}{c}{--} & 0.4083 \\ 

\multicolumn{1}{|c|}{} & $A^{(2)}$ & \multicolumn{1}{c}{29} & \multicolumn{1}{c}{2} & \multicolumn{1}{c}{0.09} & \multicolumn{1}{c}{0.06} & 0.3450 & \multicolumn{1}{c}{42} & \multicolumn{1}{c}{2} & \multicolumn{1}{c}{0.11} & \multicolumn{1}{c}{0.07} & 0.3509 & \multicolumn{1}{c}{36} & \multicolumn{1}{c}{2} & \multicolumn{1}{c}{0.10} & \multicolumn{1}{c}{0.07} & 0.3437 \\ \hline
\end{tabular}}
\end{table}

According to the results shown in Table~\ref{tab:w2v_results}, for the Hult-2003 dataset, the vectors having $d=300$ dimensions yielded the highest accuracy rates in most cases. Considering $d=300$, the most important results were reached when the percentage of embedding edge insertion was low (less than 10\%). However, for dimensions greater than $300$, values of $P$ between 0\% to 26\% yielded  high accuracy rates. Conversely, there are some exceptions where percentages of virtual edges larger than $50\%$ yielded the best performance. Regarding the parameter $w$, in most cases, the best results are reached when the parameter $w=3$ is considered. 

In the case of the Marujo-2012 dataset, Table~\ref{tab:w2v_results} shows that, generally, $d=100$ dimensions are the optimal size for the word vectors. We also observed that the typical optimal percentage of addition of embeddings type edges did not exceed $7\%$. However, there are some exceptions when high values of $P$ lead to a higher accuracy.
However, for these cases ($B^{(w)}$ and $C^{(w)}$ metrics), the addition of edges does not outperform the results obtained with the respective unweighted version of these metrics. Once again the largest context size typically achieved the best performances for the Marujo-2012 dataset. The weighted version of the PageRank ($\pi^{(w)}$) obtained the highest accuracy rate (with $k=100$, $w=3$, and 0\% of insertion of embedding edges).    

Table~\ref{tab:w2v_results} also revealed that $d=100$ and $d=300$ dimensions of the word vectors achieved the best accuracy rates for the SemEval-2010 dataset. The edge addition percentages that achieved the best results were higher compared to previous datasets. Such percentages included values ranging between 0 and $45\%$. However, for the closeness metric ($C^{(w)}$), the optimal value of $P$ was even higher, reaching 64\%. Despite this higher level of embedding enrichment, this metric obtained a poor performance when compared to all other results.  
Concerning the context parameter for co-occurrence links, both $w=1$ and $w=3$ achieved the highest accuracy rates. For the SemEval dataset, the closeness measurement ($C$) reached the best performance (with $d=300$, $w=3$, and $P=35\%$).     

In conclusion, we found that $d \leq 300$ yields competitive performance for the considered datasets. When larger values of $d$ were considered, the accuracy rates did not significantly improve. We also observed that the performance can be improved in several scenarios when larger window length and/or the inclusion of virtual edges are considered. 

\subsection{System performance analysis using the BERT model}\label{sec:results:bert}

In this section, we discuss the results we obtained considering the word vectors produced by the BERT model~\cite{devlin2018bert}. 
Because in this model each occurrence of the same word is represented by different vectors, we had to adapt our methodology concerning the insertion of virtual edges. We adopted two approaches to compute the similarity between two words. In the first approach ($BERTSim_1$), the word is represented by averaging the corresponding vector observed in each occurrence. In the second approach ($BERTSim_2$), the similarity $\textrm{sim}(a,b)$ between nodes $a$ and $b$ is computed as:
\begin{equation}
    \textrm{sim}(a,b) = \frac{1}{f_a f_b }\sum_k \sum_l \cos( v_k^{(a)}, v_l^{(b)} ),   
\end{equation}
where $v_k^{(a)}$ is the $k$-th vector representation of word $a$, $\cos$ is the cosine similarity and $f_a$ is the frequency (i.e. number of occurrences) of $a$. 
The results obtained for both approaches are depicted in Table~\ref{tab:bert_results}. The results are shown in terms of $w$ and $P$.

\begin{table}[]
\caption{Performance based on the BERT model for edge embedding creation. Acc. represents the highest accuracy rate for the considered set of parameters. For this analysis, we experimented with different values of window length and fraction of included virtual edges.}
\label{tab:bert_results}
{\footnotesize
\begin{tabular}{cc|ccccc|ccccc|}
\cline{3-12} & & \multicolumn{5}{c|}{\textbf{$BERTSim_1$}} & \multicolumn{5}{c|}{\textbf{$BERTSim_2$}}  \\ \hline

\multicolumn{1}{|c|}{\textbf{Dataset}} & \textbf{Meas.} & \multicolumn{1}{c}{\textbf{P}} & \multicolumn{1}{c}{\textbf{w}} & \multicolumn{1}{c}{\textbf{$\Gamma_1$}} & \multicolumn{1}{c}{\textbf{$\Gamma_2$}} & \textbf{Acc.} & \multicolumn{1}{c}{\textbf{P}} & \multicolumn{1}{c}{\textbf{w}} & \multicolumn{1}{c}{\textbf{$\Gamma_1$}} & \multicolumn{1}{c}{\textbf{$\Gamma_2$}} & \textbf{Acc.}  \\ \hline

\multicolumn{1}{|c|}{\multirow{12}{*}{Hult-2003}} & $k$ & \multicolumn{1}{c}{2} & \multicolumn{1}{c}{3} & \multicolumn{1}{c}{0.02} & \multicolumn{1}{c}{--} & 0.5163 & \multicolumn{1}{c}{0} & \multicolumn{1}{c}{3} & \multicolumn{1}{c}{0.03} & \multicolumn{1}{c}{--} & 0.5199 \\ 

\multicolumn{1}{|c|}{} & $s$ & \multicolumn{1}{c}{0} & \multicolumn{1}{c}{2} & \multicolumn{1}{c}{0.04} & \multicolumn{1}{c}{--} & 0.5069 & \multicolumn{1}{c}{0} & \multicolumn{1}{c}{3} & \multicolumn{1}{c}{0.04} & \multicolumn{1}{c}{--} & 0.5033  \\ 

\multicolumn{1}{|c|}{} & $\pi$ & \multicolumn{1}{c}{0} & \multicolumn{1}{c}{3} & \multicolumn{1}{c}{0.03} & \multicolumn{1}{c}{--} & 0.5235 & \multicolumn{1}{c}{0} & \multicolumn{1}{c}{3} & \multicolumn{1}{c}{0.03} & \multicolumn{1}{c}{--} & 0.5224  \\ 

\multicolumn{1}{|c|}{} & $\pi^{(w)}$ & \multicolumn{1}{c}{0} & \multicolumn{1}{c}{3} & \multicolumn{1}{c}{0.04} & \multicolumn{1}{c}{--} & 0.5080 & \multicolumn{1}{c}{0} & \multicolumn{1}{c}{3} & \multicolumn{1}{c}{0.04} & \multicolumn{1}{c}{--} & 0.5065  \\ 

\multicolumn{1}{|c|}{} & $B$ & \multicolumn{1}{c}{3} & \multicolumn{1}{c}{3} & \multicolumn{1}{c}{0.03} & \multicolumn{1}{c}{--} & 0.4973 & \multicolumn{1}{c}{0} & \multicolumn{1}{c}{3} & \multicolumn{1}{c}{0.03} & \multicolumn{1}{c}{--} & 0.4957  \\ 

\multicolumn{1}{|c|}{} & $B^{(w)}$ & \multicolumn{1}{c}{9} & \multicolumn{1}{c}{2} & \multicolumn{1}{c}{--} & \multicolumn{1}{c}{--} & 0.4727 & \multicolumn{1}{c}{0} & \multicolumn{1}{c}{1} & \multicolumn{1}{c}{--} & \multicolumn{1}{c}{--} & 0.4755  \\ 

\multicolumn{1}{|c|}{} & $C$ & \multicolumn{1}{c}{6} & \multicolumn{1}{c}{3} & \multicolumn{1}{c}{0.04} & \multicolumn{1}{c}{0.01} & 0.5016 & \multicolumn{1}{c}{0} & \multicolumn{1}{c}{3} & \multicolumn{1}{c}{0.04} & \multicolumn{1}{c}{--} & 0.4984  \\ 

\multicolumn{1}{|c|}{} & $C^{(w)}$ & \multicolumn{1}{c}{84} & \multicolumn{1}{c}{1} & \multicolumn{1}{c}{0.01} & \multicolumn{1}{c}{0.01} & 0.4737 & \multicolumn{1}{c}{3} & \multicolumn{1}{c}{1} & \multicolumn{1}{c}{0.01} & \multicolumn{1}{c}{0.01} & 0.4722  \\ 

\multicolumn{1}{|c|}{} & $EV$ & \multicolumn{1}{c}{5} & \multicolumn{1}{c}{3} & \multicolumn{1}{c}{0.08} & \multicolumn{1}{c}{0.01} & 0.5097 & \multicolumn{1}{c}{1} & \multicolumn{1}{c}{3} & \multicolumn{1}{c}{0.07} & \multicolumn{1}{c}{--} & 0.5057  \\ 

\multicolumn{1}{|c|}{} & $EV^{(w)}$ & \multicolumn{1}{c}{2} & \multicolumn{1}{c}{3} & \multicolumn{1}{c}{0.06} & \multicolumn{1}{c}{0.01} & 0.4945 & \multicolumn{1}{c}{0} & \multicolumn{1}{c}{3} & \multicolumn{1}{c}{0.05} & \multicolumn{1}{c}{--} & 0.4839  \\ 

\multicolumn{1}{|c|}{} & $A^{(1)}$ & \multicolumn{1}{c}{0} & \multicolumn{1}{c}{3} & \multicolumn{1}{c}{0.03} & \multicolumn{1}{c}{--} & 0.5318 & \multicolumn{1}{c}{0} & \multicolumn{1}{c}{3} & \multicolumn{1}{c}{0.03} & \multicolumn{1}{c}{--} & 0.5328 \\ 

\multicolumn{1}{|c|}{} & $A^{(2)}$ & \multicolumn{1}{c}{5} & \multicolumn{1}{c}{1} & \multicolumn{1}{c}{0.02} & \multicolumn{1}{c}{0.01} & 0.4859 & \multicolumn{1}{c}{0} & \multicolumn{1}{c}{3} & \multicolumn{1}{c}{0.02} & \multicolumn{1}{c}{--} & 0.4854 \\ \hline

\multicolumn{1}{|c|}{\multirow{12}{*}{Marujo-2012}} & $k$ & \multicolumn{1}{c}{7} & \multicolumn{1}{c}{2} & \multicolumn{1}{c}{0.01} & \multicolumn{1}{c}{--} & 0.5256 & \multicolumn{1}{c}{2} & \multicolumn{1}{c}{3} & \multicolumn{1}{c}{0.01} & \multicolumn{1}{c}{0.01} & 0.5246 \\

\multicolumn{1}{|c|}{} & $s$ & \multicolumn{1}{c}{2} & \multicolumn{1}{c}{2} & \multicolumn{1}{c}{0.01} & \multicolumn{1}{c}{--} & 0.5208 & \multicolumn{1}{c}{0} & \multicolumn{1}{c}{3} & \multicolumn{1}{c}{0.01} & \multicolumn{1}{c}{--} & 0.5201 \\  

\multicolumn{1}{|c|}{} & $\pi$ & \multicolumn{1}{c}{2} & \multicolumn{1}{c}{2} & \multicolumn{1}{c}{0.01} & \multicolumn{1}{c}{--} & 0.5232 & \multicolumn{1}{c}{3} & \multicolumn{1}{c}{2} & \multicolumn{1}{c}{0.01} & \multicolumn{1}{c}{--} & 0.5227  \\ 

\multicolumn{1}{|c|}{} & $\pi^{(w)}$ & \multicolumn{1}{c}{0} & \multicolumn{1}{c}{2} & \multicolumn{1}{c}{0.01} & \multicolumn{1}{c}{--} & 0.5183 & \multicolumn{1}{c}{0} & \multicolumn{1}{c}{2} & \multicolumn{1}{c}{0.02} & \multicolumn{1}{c}{--} & 0.5179  \\ 

\multicolumn{1}{|c|}{} & $B$ & \multicolumn{1}{c}{0} & \multicolumn{1}{c}{3} & \multicolumn{1}{c}{0.02} & \multicolumn{1}{c}{--} & 0.5174 & \multicolumn{1}{c}{0} & \multicolumn{1}{c}{3} & \multicolumn{1}{c}{0.02} & \multicolumn{1}{c}{--} & 0.5173  \\ 

\multicolumn{1}{|c|}{} & $B^{(w)}$ & \multicolumn{1}{c}{67} & \multicolumn{1}{c}{2} & \multicolumn{1}{c}{0.01} & \multicolumn{1}{c}{0.01} & 0.5095 & \multicolumn{1}{c}{67} & \multicolumn{1}{c}{2} & \multicolumn{1}{c}{0.01} & \multicolumn{1}{c}{0.01} & 0.5081  \\ 

\multicolumn{1}{|c|}{} & $C$ & \multicolumn{1}{c}{5} & \multicolumn{1}{c}{3} & \multicolumn{1}{c}{0.03} & \multicolumn{1}{c}{0.01} & 0.5146 & \multicolumn{1}{c}{7} & \multicolumn{1}{c}{3} & \multicolumn{1}{c}{0.03} & \multicolumn{1}{c}{--} & 0.5126  \\ 

\multicolumn{1}{|c|}{} & $C^{(w)}$ & \multicolumn{1}{c}{86} & \multicolumn{1}{c}{2} & \multicolumn{1}{c}{0.02} & \multicolumn{1}{c}{0.02} & 0.5066 & \multicolumn{1}{c}{72} & \multicolumn{1}{c}{2} & \multicolumn{1}{c}{0.01} & \multicolumn{1}{c}{--} & 0.5027  \\ 

\multicolumn{1}{|c|}{} & $EV$ & \multicolumn{1}{c}{11} & \multicolumn{1}{c}{3} & \multicolumn{1}{c}{0.04} & \multicolumn{1}{c}{0.01} & 0.5177 & \multicolumn{1}{c}{9} & \multicolumn{1}{c}{3} & \multicolumn{1}{c}{0.04} & \multicolumn{1}{c}{--} & 0.5147  \\ 

\multicolumn{1}{|c|}{} & $EV^{(w)}$ & \multicolumn{1}{c}{3} & \multicolumn{1}{c}{3} & \multicolumn{1}{c}{0.04} & \multicolumn{1}{c}{--} & 0.5096 & \multicolumn{1}{c}{2} & \multicolumn{1}{c}{3} & \multicolumn{1}{c}{0.03} & \multicolumn{1}{c}{--} & 0.5061  \\ 

\multicolumn{1}{|c|}{} & $A^{(1)}$ & \multicolumn{1}{c}{2} & \multicolumn{1}{c}{2} & \multicolumn{1}{c}{0.01} & \multicolumn{1}{c}{--} & 0.5223 & \multicolumn{1}{c}{4} & \multicolumn{1}{c}{3} & \multicolumn{1}{c}{0.01} & \multicolumn{1}{c}{--} & 0.5224 \\ 

\multicolumn{1}{|c|}{} & $A^{(2)}$ & \multicolumn{1}{c}{8} & \multicolumn{1}{c}{3} & \multicolumn{1}{c}{0.01} & \multicolumn{1}{c}{0.01} & 0.5043 & \multicolumn{1}{c}{2} & \multicolumn{1}{c}{2} & \multicolumn{1}{c}{0.01} & \multicolumn{1}{c}{--} & 0.5032 \\ \hline

\multicolumn{1}{|c|}{\multirow{12}{*}{SemEval-2010}} & $k$ & \multicolumn{1}{c}{1} & \multicolumn{1}{c}{3} & \multicolumn{1}{c}{0.09} & \multicolumn{1}{c}{--} & 0.4141 & \multicolumn{1}{c}{8} & \multicolumn{1}{c}{3} & \multicolumn{1}{c}{0.09} & \multicolumn{1}{c}{--} & 0.4137 \\

\multicolumn{1}{|c|}{} & $s$ & \multicolumn{1}{c}{0} & \multicolumn{1}{c}{3} & \multicolumn{1}{c}{0.07} & \multicolumn{1}{c}{--} & 0.4023 & \multicolumn{1}{c}{2} & \multicolumn{1}{c}{3} & \multicolumn{1}{c}{0.07} & \multicolumn{1}{c}{--} & 0.4087  \\  

\multicolumn{1}{|c|}{} & $\pi$ & \multicolumn{1}{c}{0} & \multicolumn{1}{c}{3} & \multicolumn{1}{c}{0.09} & \multicolumn{1}{c}{--} & 0.4177 & \multicolumn{1}{c}{0} & \multicolumn{1}{c}{3} & \multicolumn{1}{c}{0.09} & \multicolumn{1}{c}{--} & 0.4177  \\ 

\multicolumn{1}{|c|}{} & $\pi^{(w)}$ & \multicolumn{1}{c}{0} & \multicolumn{1}{c}{3} & \multicolumn{1}{c}{0.08} & \multicolumn{1}{c}{--} & 0.4103 & \multicolumn{1}{c}{1} & \multicolumn{1}{c}{3} & \multicolumn{1}{c}{0.09} & \multicolumn{1}{c}{--} & 0.4208  \\ 

\multicolumn{1}{|c|}{} & $B$ & \multicolumn{1}{c}{3} & \multicolumn{1}{c}{3} & \multicolumn{1}{c}{0.07} & \multicolumn{1}{c}{--} & 0.3989 & \multicolumn{1}{c}{7} & \multicolumn{1}{c}{3} & \multicolumn{1}{c}{0.08} & \multicolumn{1}{c}{0.01} & 0.4012  \\ 

\multicolumn{1}{|c|}{} & $B^{(w)}$ & \multicolumn{1}{c}{63} & \multicolumn{1}{c}{1} & \multicolumn{1}{c}{0.03} & \multicolumn{1}{c}{0.03} & 0.2436 & \multicolumn{1}{c}{4} & \multicolumn{1}{c}{1} & \multicolumn{1}{c}{--} & \multicolumn{1}{c}{--} & 0.2759  \\ 

\multicolumn{1}{|c|}{} & $C$ & \multicolumn{1}{c}{1} & \multicolumn{1}{c}{3} & \multicolumn{1}{c}{0.10} & \multicolumn{1}{c}{0.01} & 0.4165 & \multicolumn{1}{c}{20} & \multicolumn{1}{c}{3} & \multicolumn{1}{c}{0.10} & \multicolumn{1}{c}{0.01} & 0.4167  \\ 

\multicolumn{1}{|c|}{} & $C^{(w)}$ & \multicolumn{1}{c}{91} & \multicolumn{1}{c}{3} & \multicolumn{1}{c}{0.28} & \multicolumn{1}{c}{0.04} & 0.1917 & \multicolumn{1}{c}{9} & \multicolumn{1}{c}{3} & \multicolumn{1}{c}{0.03} & \multicolumn{1}{c}{--} & 0.2052  \\ 

\multicolumn{1}{|c|}{} & $EV$ & \multicolumn{1}{c}{1} & \multicolumn{1}{c}{3} & \multicolumn{1}{c}{0.09} & \multicolumn{1}{c}{--} & 0.4022 & \multicolumn{1}{c}{0} & \multicolumn{1}{c}{3} & \multicolumn{1}{c}{0.08} & \multicolumn{1}{c}{--} & 0.4007  \\ 

\multicolumn{1}{|c|}{} & $EV^{(w)}$ & \multicolumn{1}{c}{0} & \multicolumn{1}{c}{3} & \multicolumn{1}{c}{0.07} & \multicolumn{1}{c}{--} & 0.3861 & \multicolumn{1}{c}{5} & \multicolumn{1}{c}{3} & \multicolumn{1}{c}{0.09} & \multicolumn{1}{c}{0.01} & 0.3962  \\ 

\multicolumn{1}{|c|}{} & $A^{(1)}$ & \multicolumn{1}{c}{2} & \multicolumn{1}{c}{1} & \multicolumn{1}{c}{0.01} & \multicolumn{1}{c}{0.01} & 0.2830 & \multicolumn{1}{c}{0} & \multicolumn{1}{c}{1} & \multicolumn{1}{c}{--} & \multicolumn{1}{c}{--} & 0.2786 \\ 

\multicolumn{1}{|c|}{} & $A^{(2)}$ & \multicolumn{1}{c}{1} & \multicolumn{1}{c}{1} & \multicolumn{1}{c}{--} & \multicolumn{1}{c}{--} & 0.0228 & \multicolumn{1}{c}{5} & \multicolumn{1}{c}{1} & \multicolumn{1}{c}{0.06} & \multicolumn{1}{c}{0.06} & 0.0237 \\ \hline

\end{tabular}}
\end{table}

Concerning the Hult-2003 dataset, the $BERTSim_1$ approach achieved the highest accuracy rates in most cases. However, for various situations, the best results are obtained without using virtual edges or when P is lower than $6\%$. Only for the weighted closeness metric, the percentage of edge addition was quite high ($84\%$). As for the window length, $w=1$ and $w=3$ generally yielded the highest performance. The accessibility metric considering one hierarchy level ($A^{(1)}$) achieved the best performance for the Hult-2003 dataset ($BERTSim_1$ approach and $w=3$)

The results revealed that the $BERTSim_1$ approach outperformed the $BERTSim_2$ approach for the Marujo-2012 dataset. We observed that the optimal percentage of edge insertion is typically lower than $15\%$, but in particular cases, it reached high values between $67\%$ and $86\%$ (for the weighted versions of betweenness and closeness metrics).  Regarding the window length, the best results were obtained for $w \geq 2$.  The node degree ($k$) centrality obtained the highest accuracy rate considering the following parameters: $BERTSim_1$ approach, $w=2$, and a percentage of $P=7\%$  for the fraction of  virtual edges.        

Unlike the other datasets, the $BERTSim_2$ approach performed slightly better than the $BERTSim_1$ method for the SemEval-2010 dataset. The optimal value of the edge addition percentage in most cases was less than $20\%$. Higher values of $P$, however were used for both weighted versions of Betweenness and Closeness. Once again, the best results were obtained with window length $w=3$. For the SemEval-2010 dataset, the PageRank metric ($\pi$) performed better than the other centrality measurements considering the $BERTSim_2$ approach and $w=3$.  

\subsection{Summary of results and discussion} \label{sec:results:summary}

Table~\ref{tab:best_results} displays the highest accuracy rates (Acc.) of Word2Vec and BERT embedding models for each dataset. We also show the values of each parameter that achieved the best performances. We considered the following parameters relevant to our research: optimal dimension $d$ of the vectors produced by Word2Vec, and the best approach (Appr.) employed to calculate the similarity between multiple vectors generated by BERT for the same word. Most importantly, the parameters affecting the network construction ($w$ and $P$) are also reported.

\begin{table}[h]
\caption{Summary of the best results based on accuracy rate (Acc.) for both Word2Vec and BERT models. \% represents the optimal fraction of included virtual edges, $d$ is the embedding dimension, and $w$ is the context size adopted to construct co-occurrence networks.}
\label{tab:best_results}
\begin{tabular}{c|ccccc|}
\cline{2-6}
 & \multicolumn{5}{c|}{\textbf{Word2Vec model}} \\ \hline
\multicolumn{1}{|c|}{\textbf{Dataset}} & \textbf{d} & \textbf{w} & \textbf{\%} & \textbf{Meas.} & \textbf{Acc.} \\ \hline

\multicolumn{1}{|c|}{Hult-2003} & 300 & 3 & 1 & $k$ & 0.5296 \\ 
\multicolumn{1}{|c|}{Marujo-2012} & 100 & 3 & 0 & $\pi^{(w)}$ & 0.5289 \\ 
\multicolumn{1}{|c|}{SemEval-2010} & 300 & 3 & 35 & $C$ & 0.4208 \\ \hline \hline

& \multicolumn{5}{c|}{\textbf{BERT model}} \\ \hline
\multicolumn{1}{|c|}{\textbf{Dataset}} & \textbf{Apr.} & \textbf{w} & \textbf{\%} & \textbf{Meas.} & \textbf{Acc.} \\ \hline

\multicolumn{1}{|c|}{Hult-2003} & $Sim_2$ & 3 & 0 & $A^{(1)}$ & 0.5328 \\ 
\multicolumn{1}{|c|}{Marujo-2012} & $Sim_1$ & 2 & 7 & $k$ & 0.5256 \\ 
\multicolumn{1}{|c|}{SemEval-2010} & $Sim_2$ & 3 & 1 & $\pi^{(w)}$ & 0.4208 \\ \hline \hline
\end{tabular}
\end{table}

All in all, our results revealed that both Word2Vec and BERT methods have a similar performance. For the Hult-2003 dataset, the BERT-based methods were slightly better than Word2Vec, while for the Marujo-2012 dataset, the Word2Vec-based methods outperformed the BERT-based methods. Conversely, for the SemEval-2010 dataset, the results of both approaches displayed similar performance. 
This result allows the use of both techniques based on the desired property of the chosen embedding method. The training of the Word2Vec model is quite fast and has been  successfully used for representing documents for different text applications. However, here we needed to detect the optimal size of the vectors. Word2Vec also generates a single vector for a word, regardless of the context of the words. Training  documents with BERT is computationally more expensive, especially for larger datasets comprising large documents. However, the main advantage of BERT is that it generates several vectors for a word according to the number of contexts in which the word is used. This fact can lead to enhanced representations and improved performance  in different datasets. 

Concerning the network creation parameters, we found the best results with Word2vec considered vectors comprising typically less than 500 dimensions.  
In the BERT approach, the two approaches proposed to handle multiple word vectors for the same concept -- namely $BERTSim_1$ and $BERTSim_2$ -- had a similar performance. However, the $BERTSim_2$ approach requires a higher computational cost, especially when analyzing large documents. 
The experiments also showed that in most cases the percentage of addition of virtual edges is typically not very high. The performance of each system considerably decreases when high percentages of addition of virtual edges are considered. 
In conclusion, we showed -- as a proof of principle -- that the combination of further window length in the co-occurrence model and virtual edges can improve the quality of the keyword detection~\cite{vega2019multi}.

\section{Conclusion}\label{sec:conclusions}

Identifying keywords is an important task in many text mining applications. 
In this paper, we addressed this problem by generating different representations for a text using co-occurrence networks.
We considered two variations of the word adjacency model: the number of words that can be connected within the same context, and the fraction of virtual edges used to connect similar words. 
For each generated network, we evaluated several centrality measurements, including a generalization of the node degree centrality considering a network dynamics~\cite{Travencolo}. 

Our results revealed that the optimal window length in the co-occurrence network is $w = 3$, while the fraction of embeddings/virtual edges yielding the best results is typically not high. 
We also observed that the node degree, PageRank, and accessibility metrics reached the highest accuracy rates for the three datasets. The unweighted versions of the traditional measurements turned out to provide  better performance than their weighted counterparts in almost all cases. 

Our results showed, as a proof of principle, that using virtual edges can improve the informativeness of co-occurrence networks for the keyword detection task. 
Given that the informativeness of the characterization can be improved in the adopted representation, we believe that the inclusion of virtual edges could be useful in other network classification scenarios, such as in name disambiguation~\cite{amancio2015topological}.  The proposed methodology could be improved by including other model components. 
For example, edges weight modeling could be improved if both co-occurrence frequency and semantic similarity are combined, for example, via linear operations. 

Another source of improvement to the model could arise if synonyms are handled before the creation of the networks. In this way, words with similar meanings would be represented by a single node, so as to avoid redundancy in the co-occurrence networks. This could be done by taking advantage of the vectors generated by BERT, for example. Finally, our approach is limited to finding unigram keywords (keywords composed of a single word). A more general approach could consider keywords comprising two or more words.

\section*{Acknowledgments}

This study was financed in part by the Coordenação de Aperfeiçoamento de Pessoal de Nível Superior - Brasil (CAPES) - Finance Code 001. Thiago C. Silva (Grant no. 308171/2019-5, 408546/2018-2) gratefully acknowledges financial support from the CNPq foundation. Diego R. Amancio acknowledges financial support from the (Grant no. 304026/2018-2, 311074/2021-9)  and FAPESP (Grant no. 20/06271-0).

\newpage

\bibliographystyle{ieeetr}
\bibliographystyle{abbrv}

\end{document}